\documentclass[conference]{IEEEtran}
\usepackage[subrefformat=parens,labelformat=parens]{subfig}
\usepackage{amssymb}
\usepackage{amsmath}
\usepackage{bm}      
\usepackage{subfig}
\usepackage{times}
\usepackage{graphicx}
\usepackage{flushend}
\usepackage{algorithm}
\usepackage{algpseudocode}

\usepackage{xcolor}

\usepackage[numbers]{natbib}
\usepackage{multicol}
\usepackage[bookmarks=true]{hyperref}
\DeclareMathOperator*{\argmax}{arg\,max}

\algdef{S}[FOR]{ForEach}[1]{\algorithmicforeach\ #1\ \algorithmicdo}
\algnewcommand\algorithmicinput{\textbf{Input:}}
\algnewcommand\INPUT{\item[\algorithmicinput]}
\algnewcommand\algorithmicoutput{\textbf{Output:}}
\algnewcommand\OUTPUT{\item[\algorithmicoutput]}

\begin{document}

\title{Kernel Taylor-Based Value Function Approximation for Continuous-State Markov Decision Processes}

\IEEEoverridecommandlockouts
\author{Junhong Xu$^{\dagger}$, Kai Yin$^{\dagger}$, Lantao Liu
\thanks{$^{\dagger}$ Authors contributed equally. }
\thanks{J. Xu and L. Liu are with 
the Luddy School of Informatics, Computing, and Engineering  at Indiana University, Bloomington, IN 47408, USA. E-mail:
{\tt\small \{xu14, lantao\}@iu.edu}.
K. Yin is with Expedia Group, Inc. E-mail:
{\tt\small kyin@expediagroup.com}. }
}

\maketitle
\IEEEpeerreviewmaketitle

\begin{abstract}
We propose a principled kernel-based policy iteration algorithm to solve the continuous-state Markov Decision Processes (MDPs).
In contrast to most decision-theoretic planning frameworks, which assume fully known state transition models, 
we design a method that eliminates such a strong assumption,  which is oftentimes extremely difficult to engineer in reality.
To achieve this, we first apply the second-order Taylor expansion of the 
value function. 
The Bellman optimality equation is then approximated by a partial differential equation, which only relies on the first and second moments of the transition model. 
By combining the kernel representation of value function, 
we then design an efficient policy iteration algorithm whose policy evaluation step can be represented as a linear system of equations characterized by a finite set of supporting states.
We have validated the proposed method through extensive simulations in both simplified and realistic planning scenarios, and 
the experiments show that our proposed approach 
leads to a much superior performance over several baseline methods. 


\end{abstract}

\section{Introduction}
Decision-making of an autonomous mobile robot moving in unstructured environments typically requires the robot to account for uncertain action (motion) outcomes, and at the same time, maximize the long-term return.
The Markov Decision Process (MDP) is an extremely useful framework for formulating such decision-theoretic planning problems~\cite{boutilier1999decision}. 
%
Since the robot is moving in a continuous space, directly employing the standard form of MDP needs a discretized representation of the robot state and action. 
For example, in practice the discretized robot states are associated with spatial tessellation~\cite{thrun2000probabilistic}, and a grid-map like representation has been widely used for robot planning problems where each grid is regarded as a discrete state; similarly, actions are simplified as transitions to traversable grids which are usually the  very few number of adjacent grids in vicinity.

\begin{figure}[t]
    \centering
    \includegraphics[width=0.98\linewidth]{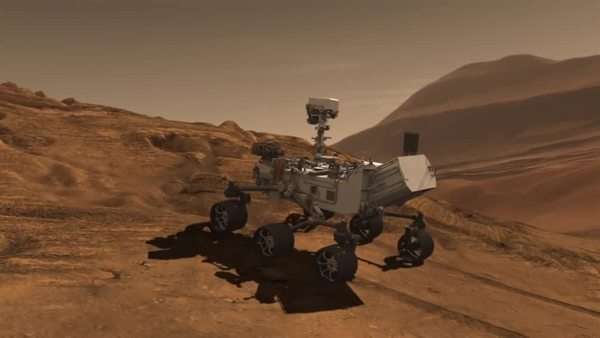}
    \caption{\small In unstructured environments, the robot needs to make motion decisions in the navigable space with spatially varying terrestrial characteristics (hills, ridges, valleys, slopes). 
    This is different from the simplified and structured environments where there are only two types of representations, i.e., either obstacle-occupied or obstacle-free. 
    Evenly tessellating the complex terrain to create a discretized state space cannot effectively characterize the underlying value function used for computing the MDP solution. 
    (Picture credit: NASA)}
    \label{fig:mars} \vspace{-10pt}
\end{figure}

However, the discretization  can be problematic. 
Specifically, if the discretization is low in resolution (i.e., large but few number of grids), the decision policy becomes a very rough approximation of the  simplified (discretized) version of the original problem; 
on the other hand, if the discretization is high in resolution, the result might be approximated well, but this will induce prohibitive computational cost and prevent the real-time decision-making. 
Finally, the characteristics of state space might be complex and it is inappropriate to conduct lattice-like tessellation which is likely to result in sub-optimal solutions. See Fig.~\ref{fig:mars} for an illustration.

Another critical issue lies in MDP's transition model
which describes the probabilistic transitions from a state to others.  
However, obtaining an accurate stochastic transition model for robot motion transition is unrealistic even without considering spatiotemporal variability. 
This is another important factor that significantly limits the applicability of MDP in many real-world problems.  
Reinforcement learning~\cite{kober2013reinforcement} does not rely on a transition model specification, but requires cumbersome training trials to learn the value function and the policy, which can be viewed as another strong 
assumption in many robotic missions. 
Thus, it is desirable that the demanding assumption of a known transition model can be relaxed. 
Fortunately, the characteristics of transition probabilistic distribution (e.g., mean, variance, or quantiles) for most robotic planning and control systems  
can be obtained from historical data or offline tests~\cite{thrun2000probabilistic}. 
If we only assume such ``partial knowledge"--mean and variance--of the transition model, we must 
re-design the modeling and solving mechanisms which will be presented in this work.
To address the above problems, we propose a kernel Taylor-based approximation approach. 
Our contributions can be summarized as follows:

\begin{itemize}
    
    
    \item 
    First,  to relax the requirement of fully known transition functions, we apply the second-order Taylor expansion to the value function~\cite{braverman2018taylor, buchli2017optimal}. The Bellman-type policy evaluation equation and Bellman optimality equation are then approximated by a partial differential equation (PDE) which only relies on the first and second moments of transition probability distribution.
    
    \item Second, 
    to improve the generalizability of the value function, 
    we use kernel functions which can represent a large number of function families for better value approximation. 
    This approximation can conveniently characterize the underlying value functions with a finite set of discrete supporting states. 

    \item 
    Finally, we develop an efficient policy iteration algorithm by integrating the kernel value function representation and the Taylor-based approximation to Bellman optimality equation. 
    The policy evaluation step can be represented as a linear system of equations with characterizing values at the finite supporting states, and the only information needed is the first and second moments of the transition function.
    This alleviates the need for heavily searching in continuous/large state space and the need for carefully modeling/engineering the transition probability.
    
\end{itemize}


\section{Related Work}
Our work primarily focuses on value function approximation in large/continuous state space using only minimum prior knowledge of the transition function.
%
A major challenge for solving the continuous-state MDP 
involves the search in a large-scale (usually infinite) state space. 
Popular methods in robotics that avoid intractability of computing the value function over the continuous space are by tessellating the continuous state space into grids~\cite{pereira2013risk, otte2016any, fu2015sense, al2013wind, baek2013optimal}. 
However, this naive approach does not scale well and may give inferior performance when the problem size increases, known as the curse of dimensionaliy~\cite{bellman2015adaptive}.
A more advanced discretization technique that alleviates this problem is by adaptive discretization~\cite{gorodetsky2015efficient, liu2018solution, Liu-RSS-19, munos2002variable, lavalle2006planning}.

Alternative methods tackle this challenge by representing and also approximating the value function by a set of basis functions or some parametric functions~\cite{bertsekas1996neuro, sutton2018reinforcement, munos2008finite}. 
The parameters can be optimized by minimizing the Bellman residual~\cite{antos2008learning}.
However, these methods are not applicable in complicated problems because defining features to approximate the value function linearly is non-trivial. 
This weakness may be resolved through kernel methods~\cite{hofmann2008kernel}.
Because the weights in linear combination of basis functions can be presented by  their product (through so-called duel form of least squares~\cite{shawe2004kernel}), the weights can be written in terms of kernel functions and value functions at supporting states. Once value functions at supporting states are obtained, the approximation to value function at any state is also determined. 
The approach to approximating the value function by kernel functions is referred as the {\em direct kernel-based method} in this paper. In addition, it is generally hard to find a suitable nonlinear function (such as neural networks) to approximate the value function ~\cite{glorot2010understanding}. 
There is a vast literature on kernelized value function approximations in reinforcement learning~\cite{engel2003bayes, kuss2004gaussian, taylor2009kernelized, xu2007kernel}, 
but few studies in robotic planning problems leveraged this approach. A recent application of work~\cite{engel2003bayes} on marine robots can be found in~\cite{martin2018sparse}.

Unfortunately, all these schemes rely on either fully known transitions in MDP or the selection of basis functions, 
which are difficult to obtain in practice. 
Therefore, the challenge becomes {\em how to design a principled methodology without explicitly relying on basis functions and without full knowledge of transitions in MDP, which will be addressed in this work}.

\section{Preliminary Material}
\subsection{Markov Decision Processes}
We formulate the robot decision-theoretic planning problem 
as an infinite horizon discounted Markov Decision Process (MDP) with continuous states and finite actions. 
An infinite horizon discounted MDP is defined by a 5-tuple $\mathcal{M}\triangleq\langle\mathbb{S}, \mathbb{A}, T, R, \gamma\rangle$,
where $\mathbb{S}=\{s\} \subseteq \mathbb{R}^{N}$ is the $n$-dimensional continuous state space and
$\mathbb{A} = \{a\}$ is a finite set of actions. $\mathbb{S}$ can be thought of as the robot workspace in our study. 
A robot transits from a state $s$ to the next $s'$ by taking an action $a$ in a stochastic environment and obtains a reward $R(s, a)$. Such transition is governed by a conditional probability distribution $T(s, a, s') \triangleq p(s'|s, a)$ which is also termed as transition model (or transition function); the reward $R(s, a)$, a mapping from a pair of state and action to a scalar value, specifies the short-term objective that the robot receives by taking action $a$ at state $s$.
The final element $\gamma \in [0, 1)$ in $\mathcal{M}$ is a discount factor which will be used in the expression of value function.

We consider the class of deterministic policies $\Pi$, which defines a mapping $\pi \in \Pi: \mathbb{S}\rightarrow \mathbb{A}$ from a state to an action. 
The expected discounted cumulative reward for any policy $\pi$ starting at any state $s$ is expressed as
\begin{equation}\label{eq:value-fn-policy}
    v^{\pi}(s) = \mathbb{E}[\sum_{k=0}^{ \infty} \gamma^{k} R(s_k, a_k)|s_0 = s, a_k = \pi(s_k)],
\end{equation}
We can rewrite the above equation recursively as follows
\begin{equation}\label{eq:value-function}
\begin{split}
    v^{\pi}(s) =& R(s, \pi(s)) + \gamma\,\mathbb{E}^{\pi}[v^{\pi}(s')|s] \triangleq \mathcal{B}^{\pi}v^{\pi}(s),
\end{split}
\end{equation}
where 
$\mathcal{B}^{\pi}$ is called the {\em Bellman operator}, and $\mathbb{E}^{\pi}[v^{\pi}(s')|s]=\int p(s'|s, \pi(s))v^{\pi}(s')\,ds'$.
The function $v^{\pi}(s)$ in Eq.~\eqref{eq:value-function} is usually called the {\em state value function} of the policy $\pi$. 
Solving an MDP is to find the optimal policy $\pi^*$ with the optimal value function which satisfies the Bellman optimality equation 
\begin{equation}\label{bellman-equation}
  v^{\pi^*}(s) = \max_{\pi}\left\{R(s, \pi(s)) + \gamma \int p(s' | s, \pi(s))v^{\pi}(s')\,ds'\right\}.
\end{equation}

\subsection{Approximate Policy Iteration via Value Function Representation}\label{sec:primal}


To solve an MDP, value iteration and policy iteration are the most prevalent approaches. It has been shown that the value iteration and policy iteration can achieve similar state-of-the-art performances in terms of solution quality and running time~\cite{bertsekas1995dynamic, sutton2018reinforcement}. 
Our work will be built upon policy iteration and here 
we provide a summary of the important value function approximation process used in the policy iteration \cite{powell2016perspectives,gordon1999approximate}. 

Policy iteration requires initialization of the policy (can be random), based on which a system of linear equations can be established  where each equation is exactly the value function (Eq.~\eqref{eq:value-function}). When the states in the MDP are finite, the solution to this linear system yields incumbent values for all states~\cite{puterman2014markov}. This step is called \textit{policy evaluation}. The second step is to improve the current policy by greedily improving local actions based on the incumbent values obtained. This step is called \textit{policy improvement}. 
Through iterating these two steps, we can find the optimal policy and a unique solution to the value function that satisfies Eq.~\eqref{eq:value-fn-policy} for every state. 

If, however, the states are continuous or the number of states is infinite, it is difficult to evaluate 
the value function at every state. One must resort to  approximate solutions. 
Suppose that the value function can be represented by a weighted linear combination of known functions where only weights are to be determined, then 
a natural way to go is leveraging the Bellman-type equation, i.e., Eq.~\eqref{eq:value-function}, to compute the weights. 
Specifically, given an arbitrary policy, the representation of value function can be evaluated at a finite number of states, leading to a linear system of equations whose solutions can be viewed as  weights~\cite{lagoudakis2003least}. 
This obtained representation of value function can be used to improve the current policy. The remaining   procedure is then similar to the standard policy iteration method.
The final obtained value function representation serves as an approximated optimal value function \textit{for the whole continuous state space}, and the corresponding policy can be obtained accordingly. 

Formally, let the value function approximation under policy $\pi$ be 
\begin{equation}\label{basis-approx}
 v^{\pi}(s)\simeq v(s; w^{\pi}) = \sum_{i=1}^m w_i^{\pi}\cdot \phi_i(s),    
\end{equation}
where $\phi_i\in\Phi\triangleq\{\phi_1, \ldots, \phi_m\}$. The set $\Phi$ is 
the \textit{basis functions} in literature~\cite{powell2016perspectives}.
A finite number of supporting states $\mathbf{s} = \{s^{1}, \ldots, s^N\}$, $N>m$ can be selected, which are minimized via the 
squared {\em Bellman error} over $\mathbf{s}$, defined by 
$\mathcal{L}(w^{\pi}) = \sum_{i=1}^{N}({v(s^i;w^{\pi}) - \mathcal{B}^{\pi}v({s}^i; w^{\pi}))^2}.$
The solution for $w^{\pi}$ may have a closed form in terms of the basis functions, transition probabilities, and rewards~\cite{lagoudakis2003least}. By policy iteration, the final solution for $v(s; w^{\pi})$ can be obtained. 

Note that $v(s; w^{\pi})$ may be generalized to any parametric nonlinear functions such as neural networks, and that the selection of supporting states $\mathbf{s}$ needs to take account of the characteristics of the underlying value functions (in our robotics decision-theoretic planning scenarios, it relates to the landscape geometry of the terrain). 

\section{Kernel Taylor-Based Approximate Policy Iteration}

Our objective is to design a principled kernel-based policy iteration approach by leveraging kernel methods to solve the continuous-state MDP. 
In contrast to 
most decision-theoretic planning frameworks which assume fully known MDP transition probabilities~\cite{boutilier1999decision, puterman2014markov}, 
we propose a method that eliminates such a strong premise which oftentimes is  extremely difficult to engineer in practice. 
To overcome this challenge, 
first we apply the second-order Taylor expansion of the kernelized value function (Section~\ref{sec:taylored-policy-eval}). 
The Bellman optimality equation is then approximated by a partial differential equation which only relies on the first and second moments of transition probabilities (Section~\ref{sec:bellman-eqn-pde}). 
Combining the kernel representation of value function, this approach efficiently tackles the continuous or large-scale state space search with  minimum prerequisite knowledge of state transition model (Sections~\ref{kernel-taylored-P-E} and ~\ref{K-T-P-I}). 
Finally, the experiments show that our proposed approach is very powerful and flexible, and reveal great advantages over several baseline approaches (Section~\ref{Experiment}).

\subsection{Taylored Approximate Policy Evaluation Equation}\label{sec:taylored-policy-eval}

To design an efficient approach for solving MDP based decision-theoretic planning problems, we essentially have two elements to deal with: the value function and the Bellman optimality equation. 
If we directly apply kernel methods to approximate the value function (referred to as the {\em direct kernel-based method}), we can avoid explicitly specifying basis functions as mentioned in Section~\ref{sec:primal}.  
But it still requires fully known MDP transition probabilities, and it needs the exact Bellman optimality equations to develop the policy iteration method.

In contrast to the direct kernel-based approach, we consider an approximation to the Bellman-type equation by using only first and second moments of transition functions.
This will allow us to obtain a nice property that a complete and accurate transition model is not necessary; instead, only the important statistics such as mean and variance (or covariance) will be sufficient. 
To better describe the basic idea, we keep our discussions on a surface-like terrain and use that  surface as the decision-theoretic planning workspace, i.e., $s=[x,y]^T\,\overset{\Delta}{=}\,[s_x, s_y]^T$, though our approaches apply to the state space of any dimensions. 

Formally, suppose that the value function $v^{\pi}(s)$ for any given policy $\pi$ has continuous first and second order derivatives. 
We subtract both hand-sides by $v^{\pi}(s)$ from Eq.~\eqref{eq:value-function} and 
then take Taylor expansions of value function around $s$ up to second order~\cite{braverman2018taylor}: 
\begin{align}
    &-R(s, \pi(s)) \nonumber\\
    &=\gamma\left(\mathbb{E}^{\pi}[v^{\pi}(s')\mid s] - v^{\pi}(s)\right) - (1-\gamma)\,v^{\pi}(s) \nonumber\\
    &=\gamma\int p(s'| s,\pi(s))(v^{\pi}(s') - v^{\pi}(s))\,ds' - (1-\gamma)\,v^{\pi}(s) \nonumber\\
    &\simeq \gamma\,\Big((\mu^{\pi}_s)^T\,\nabla v^{\pi}(s) +\frac{1}{2}\nabla\cdot \sigma_s^{\pi}\nabla v^{\pi}(s)\Big) - (1-\gamma)\,v^{\pi}(s), \label{bellman-type-pde}
\end{align}
where $\mu^{\pi}_s$ and $\sigma^{\pi}_s$ are the first moment (i.e., mean, a 2-dimensional vector) and the second moment (i.e., covariance, a 2-by-2 matrix) of transition functions, respectively, with the following form
\begin{subequations}\label{mu-sigma-eqns}
\begin{align}
    (\mu^{\pi}_s)_i &= \int p(s'| s,\pi(s))(s'_i - s_i)\,ds', \\
    (\sigma^{\pi}_s)_{i, j} &= \int p(s' | s,\pi(s))(s'_i - s_i)(s'_j - s_j)\,ds',
\end{align}
\end{subequations}
for $i, j\in\{x, y\}$; the operator $\nabla \overset{\Delta}{=} (\partial/\partial x, \partial/\partial y)$ and the notation $\cdot$ in the last equation indicate an inner product. 
To be clear, we present the expression for the following operator 
\begin{eqnarray*}
    \nabla\cdot \sigma^{\pi}_s\nabla = \sigma^{\pi}_{xx}\frac{\partial^2}{\partial x^2} +\sigma^{\pi}_{xy}\frac{\partial^2}{\partial x\partial y}
    +\sigma^{\pi}_{yx}\frac{\partial^2}{\partial y\partial x}
    +\sigma^{\pi}_{yy}\frac{\partial^2}{\partial y^2}.
\end{eqnarray*}

Since Eq. (\ref{bellman-type-pde}) approximates calculation of Eq.~\eqref{eq:value-function} in the policy evaluation stage, the solution to Eq. (\ref{bellman-type-pde}) thus provides the value function approximation under current policy $\pi$. 
Eq.~\eqref{bellman-type-pde} also implies that we only need to use the mean $(\mu^{\pi}_s)_i$ and convariance $(\sigma^{\pi}_s)_{i, j}$ instead of the original transition model $p(s'| s,\pi(s))$ to approximate the value function.

\subsection{Approximate Bellman Optimality Equation via PDE}\label{sec:bellman-eqn-pde}
We need to analyze the necessary boundary conditions to Eq.~\eqref{bellman-type-pde} which is a partial differential equation (PDE), and develop an approximation methodology to the Bellman optimality equation which is the foundation for efficient MDP solution.
To achieve these, first, the directional derivative of the value function with respect to the unit vector normal at the boundary states 
must be zero. 
(Note, the value function should not have values on obstacles or outside 
the state space $\mathbb{S}$.)
Second, in order to ensure a unique solution, we constrain the value function at the goal state to a fixed value. 

Let us denote the boundary of entire continuous planning region/workspace by $\partial\mathbb{S}$ and the goal state by $s_g$. Suppose the value function at $s_g$ is $v_{g}$.
Section~\ref{sec:taylored-policy-eval} implies that the Bellman optimality equation Eq.~\eqref{bellman-equation} can be approximated by the following PDE:
\begin{eqnarray}
 0&=&\max_{\pi}\Big\{\gamma\,\Big((\mu_s^\pi)^T\nabla v^{\pi}(s) +\frac{1}{2}\nabla\cdot \sigma_s^\pi\nabla v^{\pi}(s)\Big) + \nonumber\\
 &{}& R(s, \pi(s)) - (1-\gamma)\,v^{\pi}(s)\Big\}, 
 \label{diffusion-pde}
\end{eqnarray}
with boundary conditions
\begin{subequations}\label{boundary-conditions}
\begin{eqnarray}
    \sigma_s^{\pi}\,\nabla v^{\pi}(s)\cdot \hat{\bm{n}} &=& 0, \mbox{  on } \partial\mathbb{S} \label{boundary-condition-a}\\
    v^{\pi}(s_g) &=& v_g,  \label{boundary-condition-b}
\end{eqnarray}
\end{subequations}
where $\hat{\bm{n}}$ denotes the unit vector normal to $\partial\mathbb{S}$ pointing outward. The condition (\ref{boundary-condition-a}) is a type of homogeneous Neumann condition, and condition (\ref{boundary-condition-b}) can be thought of as a Dirichlet condition in literature \cite{Evens2010}.
This elegantly approximates the classic Bellman optimality equation by a convenient PDE representation.
In the next section, we will leverage the kernelized representation of the value function to avoid difficulties of 
directly solving PDE. The kernel method will help transform the problem to a linear system of equations with 
unknown values at the finite supporting states.


\subsection{Kernel Taylor-Based Approximate Policy Evaluation}\label{kernel-taylored-P-E}
With aforementioned formulations, another critical research question is whether the value function can be represented by some special functions that are able to approximate large function families in a convenient way.
We tackle this question by using a kernel method to represent the value function. 
Thanks to Eq.~\eqref{diffusion-pde} which allows us to extend with kernelized policy evaluation for Taylored value function approximation. 

Specifically, let $k(\cdot, \cdot)$ be a generic kernel function$^{\dagger}$~\cite{hofmann2008kernel}.
For a set of selected finite supporting states $\mathbf{s}=\{s^{1}, \ldots, s^N\}$, let $\mathbf{K}$ be the Gram matrix with $[\mathbf{K}]_{i,j}=k(s^i, s^j)$, and $\mathbf{k}(\cdot, \mathbf{s})=[k(\cdot, s^1), \ldots, k(\cdot, s^N)]^T$. Given a policy $\pi$, assume the value functions at $\mathbf{s}$ are   $V^{\pi}=[v^{\pi}(s^1), \ldots, v^{\pi}(s^N)]^T$. Then, for any state $s'$, the kernelized value function has the following form
\begin{equation}\label{kernerlized-value-func}
    v^{\pi}(s') = \mathbf{k}(s', \mathbf{s})^T\,\left(\lambda\mathbf{I}+\mathbf{K}\right)^{-1}\,V^{\pi},
\end{equation}
where $\lambda \geq 0$ is a {\em regularization factor}. When $\lambda = 0$, it links to the kernel ordinary least squares estimation of $w^{\pi}$ in Eq.~\eqref{basis-approx}; when $\lambda > 0$, it refers to the ridge-type regularized kernel least squares estimation~\cite{shawe2004kernel}. Furthermore, Eq.~\eqref{kernerlized-value-func} implies that as long as the values $V^{\pi}$ are available, the value function for any state can be immediately obtained. Now our objective is to get $V^{\pi}$ through Eq.~\eqref{bellman-type-pde} and boundary conditions Eq.~\eqref{boundary-conditions}.

\footnotetext{$^{\dagger}$ It is worth mentioning that our approach of utilizing kernel methods is to approximate the function. This usage should be distinguished from that in the machine learning literature where kernel methods are used to learn patterns from data.}

Plugging the kernelized value function representation into Eq.~\eqref{bellman-type-pde}, we end up with the following linear system:
\begin{equation}\label{linear-system-V}
\left(\mathbf{M}^{\pi} \left(\lambda\mathbf{I}+\mathbf{K}\right)^{-1} -(1-\gamma)\,\mathbf{I}\right)\,V^{\pi} = \mathbf{R}^{\pi},
\end{equation}
where $\mathbf{I}$ is an identity matrix, $\mathbf{R}^{\pi}$ is a $N\times 1$ vector with element $[\mathbf{R}^{\pi}]_i = -R(s^i, \pi(s^i))$, and $\mathbf{M}^{\pi}$ is a matrix whose elements are:
\begin{equation}\label{equations-for-M}
[\mathbf{M}^{\pi}]_{i,j} = \gamma\left ((\mu^{\pi}_{s^i})^T\nabla_{s^i} + \frac{1}{2}\nabla_{s^i}\cdot \sigma_{s^i}^{\pi}\nabla_{s^i} \right)k(s^i, s^j).
\end{equation}
Note that $\nabla_{s^i}$ indicates the derivatives with respect to $s^i$, i.e., $\nabla_{s^i}\overset{\Delta}{=} (\partial/\partial s^i_x, \partial/\partial s^i_y)$. 
In Appendix, 
we provide a concrete example using Gaussian kernels which lead to closed-form expressions and  
are widely utilized in practice. 

The solutions to the system Eq.~\eqref{linear-system-V} yield values of $V^{\pi}$. 
These values further allow us to obtain the value function~\eqref{kernerlized-value-func} for any state under current policy $\pi$. 
This completes modeling our kernel Taylor-based approximate policy evaluation framework.

\subsection{Kernel Taylor-Based Approximate Policy Iteration}\label{K-T-P-I}

\begin{algorithm}[t] 
\caption{Kernel Taylor-Based Approximate Policy Iteration}
\label{alg:kernel-based-policy-iteration}
\begin{algorithmic}[1]
    \INPUT{
    A set of supporting states $\mathbf{s} = \{\mathbf{s}^1, ..., \mathbf{s}^N\}$;
    the kernel function $k(\cdot, \cdot)$; the regularization factor $\lambda$; 
    the MDP $\langle \mathbb{S}, \mathbb{A}, T, R, \gamma \rangle$. 
    }
    \OUTPUT{The kernelized value function Eq.~\eqref{kernerlized-value-func} for every state and corresponding policy.}
    \State Initialize the action at the supporting states.
    \State Compute the matrix $\mathbf{K} + \lambda \mathbf{I}$ and its inverse. 
    \Repeat
        \State // Policy evaluation step
        \State Solve for $V^{\pi}$ according to Eq.~\eqref{linear-system-V} in Section~\ref{sec:taylored-policy-eval}. 
        \State // Policy improvement step
        \For{$i = 1, ..., N$}
            \State Update the action at the supporting state $s^i$ based on Eq.~\eqref{eq:policy-improvement}. 
        \EndFor
    \Until actions at the supporting states do not change. 
  \end{algorithmic}
\end{algorithm} 

\begin{figure*}[t] \vspace{-5pt}
    \centering
    \subfloat[Kernel Taylor-based PI]{\label{fig:taylred-pi-10-10}\includegraphics[width=0.225\linewidth]{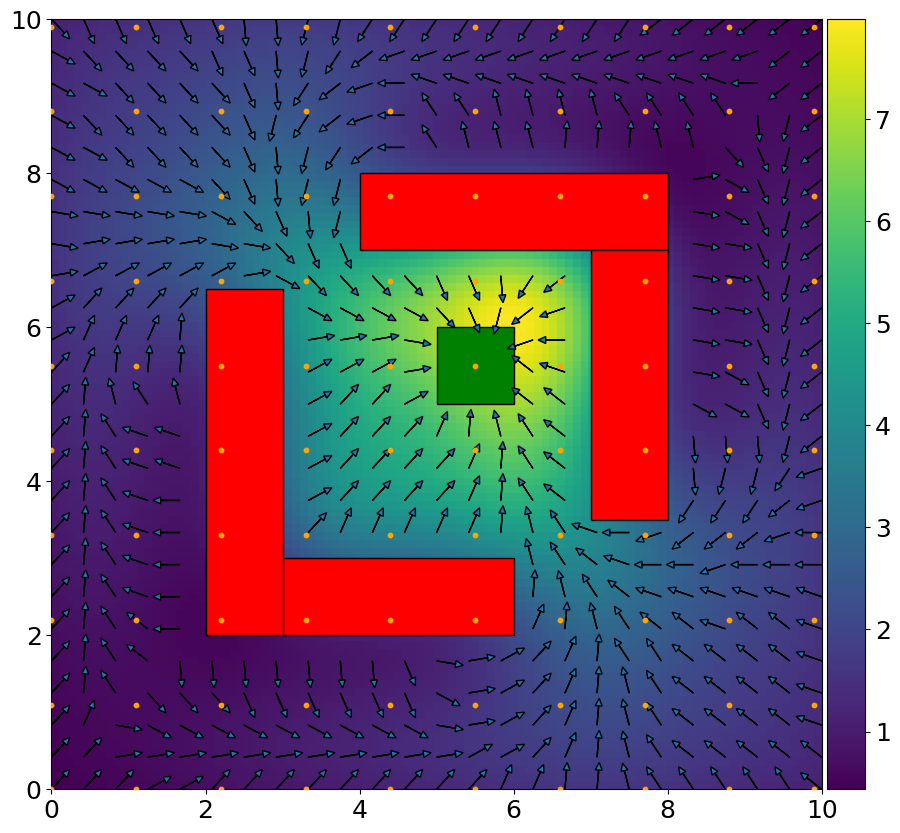}}
    \subfloat[Direct kernel-based PI]{\label{fig:grid-pi-10x10}\includegraphics[width=0.23\linewidth]{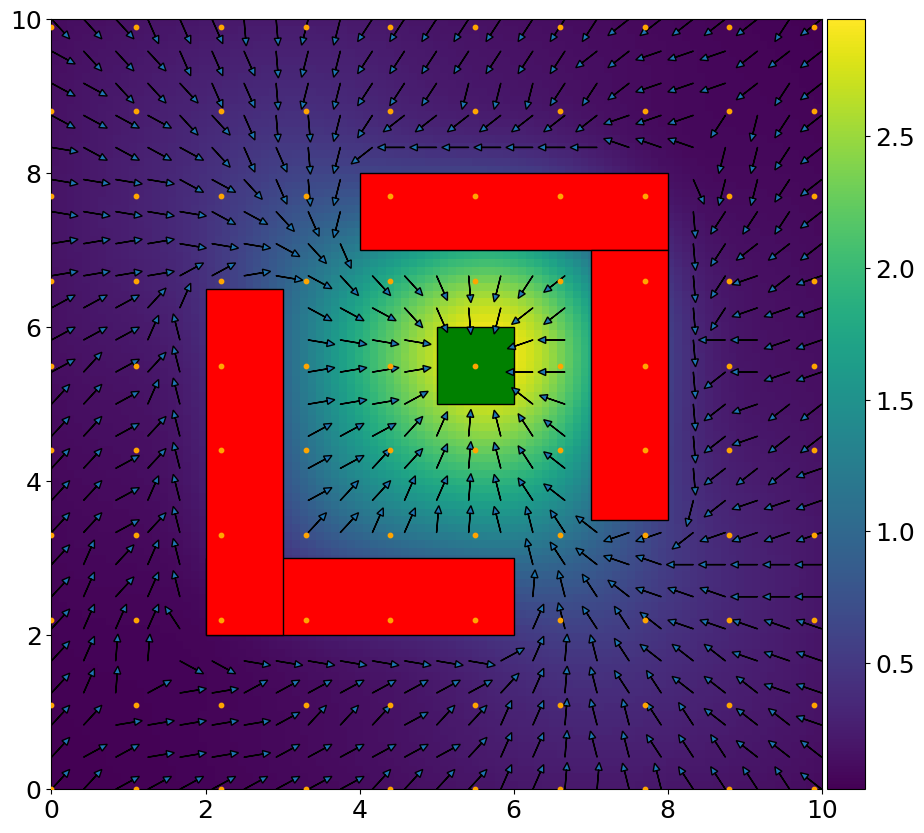}}
    \subfloat[NN]{\label{fig:grid-pi-10x10}\includegraphics[width=0.23\linewidth]{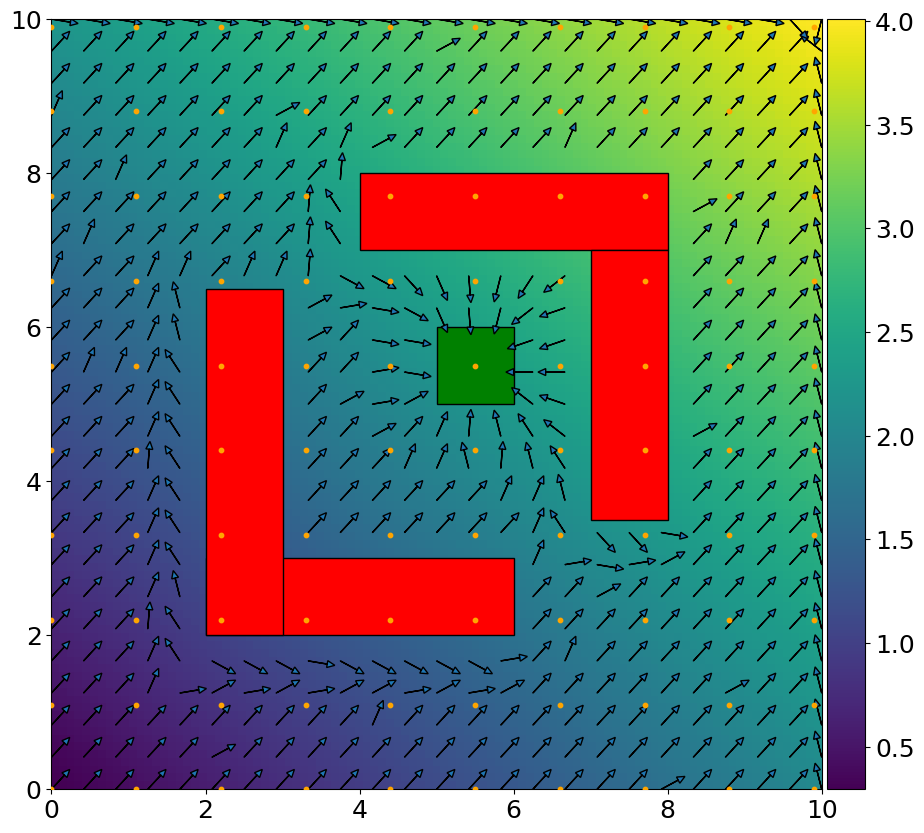}}
    \subfloat[Grid-based PI]{\label{fig:taylred-pi-10-10}\includegraphics[width=0.23\linewidth]{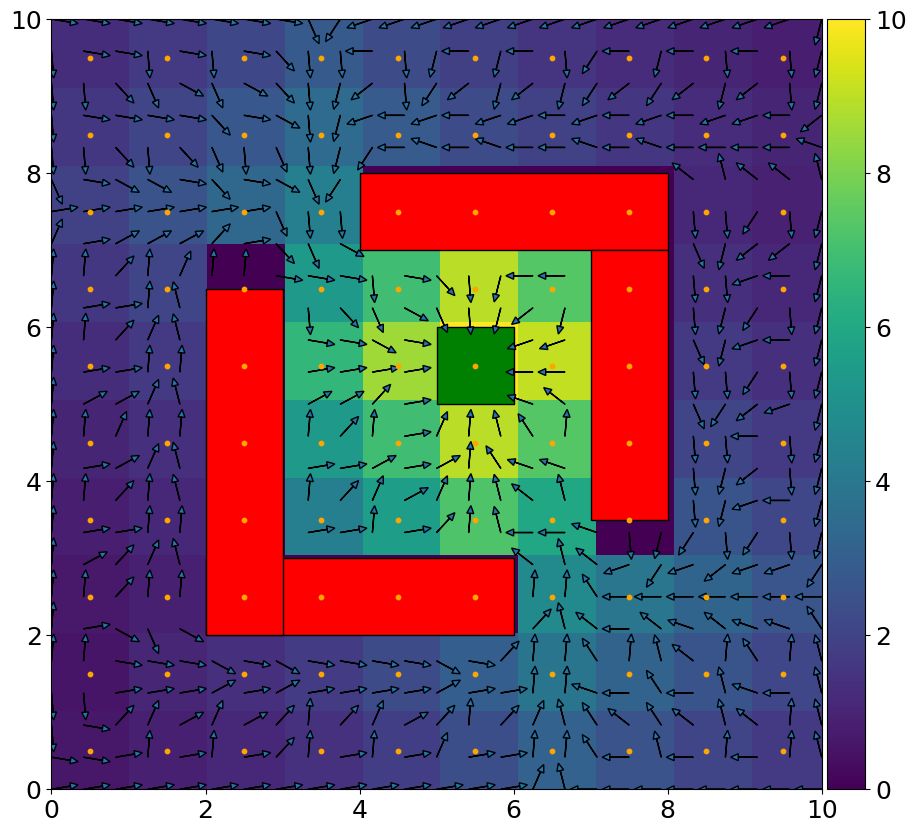}}
    \caption
    {\small
        Evaluation with a traditional simplified scenario where obstacles and goal are depicted as red and green blocks, respectively. 
        We compare the final value function and the final policy obtained from (a) kernel Taylor-based PI, (b) direct kernel-based PI, (c) NN, and (d) grid-based PI. 
        A brighter background color represents a higher state value.
        The policies are the arrows (vector fields), and each arrow points to some next waypoint.
        Orange dots denote 
        supporting states or the grid centers (in the case of grid-based PI).
    }
    \label{fig:value-fn-policy}
\end{figure*}





With the above new model, our next step is to design an implementable algorithm that can solve the continuous-state MDP efficiently.
We extend the classic policy iteration mechanism which iterates between the policy evaluation step and the policy improvement step until convergence to find the optimal policy as well as its corresponding optimal value function.

Because our kernelized value function representation depends on the finite supporting states $\mathbf{s}$ instead of the whole state space, we only need to improve the policy on $\mathbf{s}$.
Therefore, the policy improvement step in the $(k+1)$-th iteration is to produce a new policy according to
{\small
\begin{align}
     \pi_{k+1}(s)=\argmax_{a\in\mathbb{A}}\Big\{R(s, a)+\gamma\Big((\mu_s^{a})^T\,\nabla+ 
     \frac{1}{2}\nabla\cdot \sigma_s^a\nabla\Big) v^{\pi_k}(s) \Big\} \label{eq:policy-improvement}
\end{align}
}
where $s\in\mathbf{s}$, $\pi_{k}$ and $\pi_{k+1}$ are the current policy and the updated policy, respectively. Note that $\mu_s^a$ and $\sigma_s^a$ depend on $a$ through the transition function $p(s'|s, a)$ in Eq.~\eqref{mu-sigma-eqns}. 
Compared with the approximated Bellman optimality equation (Eq.~\eqref{diffusion-pde}), Eq.~\eqref{eq:policy-improvement} drops the term $(1-\gamma)v^{\pi_k}(s)$. This is because $v^{\pi_k}(s)$ does not explicitly depend on action $a$.
The value function of the updated policy satisfies $v^{\pi_{k+1}}(s) \geq v^{\pi_{k}}(s)$~\cite{bertsekas1995dynamic}.
If the equality holds, the iteration converges.
%


The final kernel Taylor-based policy iteration algorithm is pseudo-coded in Alg.~\ref{alg:kernel-based-policy-iteration}.
It first initializes the actions at the finite supporting states and then iterates between policy evaluation and policy improvement.
Since the supporting states as well as the kernel parameters do not change, the regularized kernel matrix and its inverse are computed only once at the beginning of the algorithm. 
This greatly reduces the computational burden caused by matrix inversion.
Furthermore, due to the finiteness of the supporting states, the entire algorithm views the policy $\pi$ 
as a table and only updates the actions at the supporting states using Eq.~\eqref{eq:policy-improvement}.
The algorithm stops and returns the supporting state values when the actions are stabilized. 
We can then use these state-values to get the final kernel value function that approximates the optimal solution. 
The corresponding policy for every continuous state can then be easily obtained from this kernel value function~\cite{si2004handbook}.

Intuitively, this proposed framework is flexible and powerful due to the following reasons: 
rather than tackling the difficulties in solving the PDE which approximates the original Bellman-type equation, we use the kernel representation to convert the problem to a system of linear equations with characterizing values at the finite discrete supporting states. 
From this viewpoint, our proposed method nicely balances the trade-off between searching in finite states and that in infinite states. In other words, our approach leverages the kernel methods and Bellman optimal conditions under the practical assumptions.



\section{Experiments} \label{Experiment}

To validate our method, we consider two mobile robot decision-theoretic planning tasks. 
The first one is a goal-oriented planning problem in a simple environment with obstacle-occupied and obstacle-free spaces. This will help us to evaluate basic algorithmic properties. 
In the second task, we demonstrate that our method can be applied to a more realistic as well as more challenging navigation scenario  on Mars surface~\cite{maurette2003mars}, where the robot needs to take the elevation of the terrain surface into account (i.e., ``obstacles" are implicit).
In both tasks, we assume that the estimates of the first two moments of the transition probability are obtained from the past field experiments.
To be concrete, in both experiments we use the Gaussian kernel given in Appendix~\ref{Gaussian-kernel}.

\begin{figure}[t] 
    \centering
    \subfloat[$6\times6$]{\label{fig:taylor-pi-36}\includegraphics[width=0.485\linewidth]{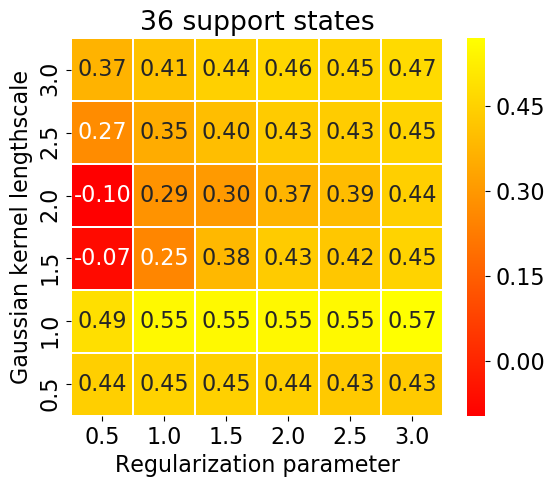}}
    \subfloat[$7\times7$]{\label{fig:taylore-pi-49}\includegraphics[width=0.5\linewidth]{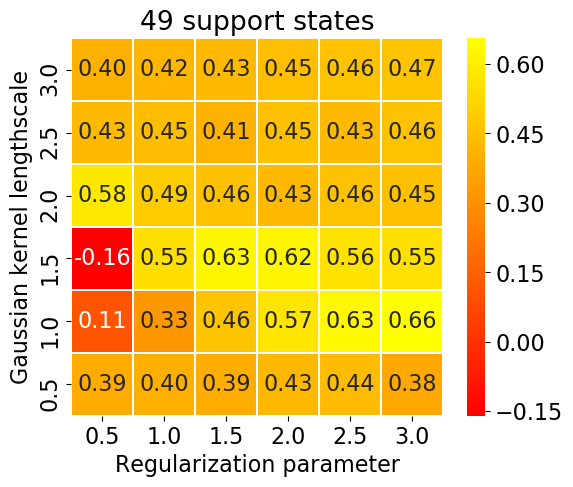}}\\
    \subfloat[$10\times10$]{\label{fig:taylor-pi-100}\includegraphics[width=0.49\linewidth]{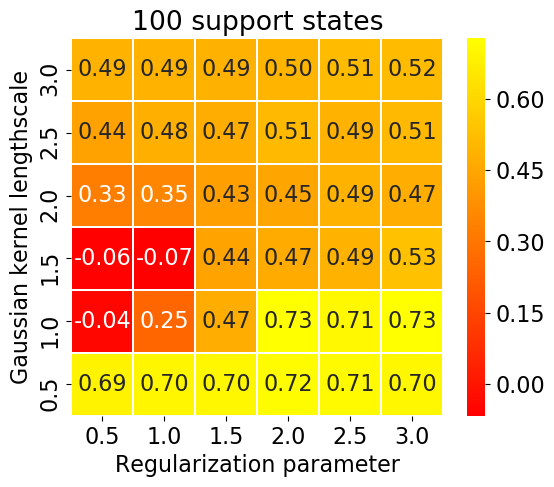}}
    \subfloat[$11\times11$]{\label{fig:taylor-pi-121}\includegraphics[width=0.49\linewidth]{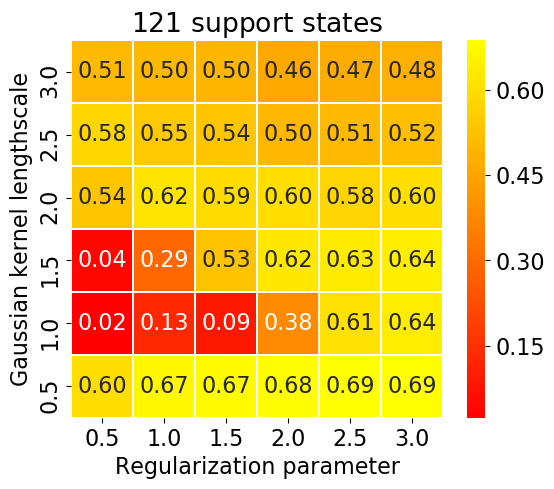}}
    \caption
    {\small 
        The performance matrix obtained by the hyperparameter search using (a) $6\times 6$; (b) $7 \times 7$; (c) $10\times 10$; and (d) $11 \times 11$ evenly-spaced supporting states. 
        Rows and columns represent different Gaussian kernel lengthscale and regularization parameters, respectively.
        The numbers in the heatmap represent the average return of the final policy obtained using the corresponding hyperparameter combination.
        The colorbar is shown on the right side of each table.
    }
    \label{fig:hyper-parameter}
    \vspace{-10pt}
\end{figure}

\subsection{Plane Navigation}\label{sec:plane-navigation}

\subsubsection{Setup} 
Our first experiment is a 2D plane navigation problem, where the obstacles and a goal area are represented in a $10m \times 10m$ environment, as shown in Fig.~\ref{fig:value-fn-policy}. 
The state space for this task is a 2-dimensional euclidean space, i.e., $ s=[s_x, s_y]^T$ and $s \in \mathbb{S} \subseteq \mathbb{R}^2$. 
The action space is a finite set of $Q$ points $\mathbb{A}(s) = \{a_i(s) | i \in \{1, ..., Q\}\}$.
Each point $a_i(s)  = [s_x + r\cos(\frac{2\pi i}{Q}), s_y+r\sin(\frac{2\pi i}{Q})]^T$ is an action generated on a circle centered at the current state with a radius $r$.
In this experiment, we set the number of actions $Q = 12$ and the action radius as $r = 0.5m$.
An action point can be viewed as the ``carrot-dangling" waypoint for the robot to follow,
which serves as the input to the low-level motion controller.
For the reward function, we set the reward of arriving at the goal and obstacle states to be $+1$ and $-1$, respectively. 
Since the reward now depends on the next state, we use Monte-carlo sampling to estimate the expectation of $R(s,a)$.
The discount factor for the reward is set to $\gamma = 0.9$.
We set the obstacle areas and the goal as  absorbing states, i.e., the robot can not transit to any other states if they are in these states.
To satisfy the boundary condition mentioned in Section~\ref{sec:bellman-eqn-pde}, we allow the robot to receive rewards at the goal state, but it cannot receive any reward if its current state is within an obstacle. 
Thus, the goal state value is $\frac{1}{1-\gamma} = 10$.


\subsubsection{Performance measure}
Since the ultimate goal of planning is to find the optimal policy, our performance measure is based on the quality 
of the policy. 
A policy is better if it achieves a higher expected cumulative reward starting from every state.
Because it is impossible to evaluate over the infinite number of states,
we numerically evaluate the quality of a policy using the average return criterion~\cite{islam2017reproducibility}. 
In detail, 
we first uniformly sample $Q=10^4$ states to ensure a thorough performance evaluation.
Then, for each sampled state, we execute the policy to generate 
multiple trajectories, where each trajectory ends when it arrives at a terminal state (goal or obstacle) or reaches an allowable maximal number of steps. 
This procedure gives us an expected performance of the policy at any state by averaging the discounted sum of rewards over all the trajectories starting from it.
Now, we can calculate the average return criterion by averaging over the performance of sampled states.
A higher value of the average return implies that, on average, the policy gives better performance over the entire state space.

\subsubsection{Results} 
\begin{figure}[t] 
    \centering
    \includegraphics[width=0.7\linewidth]{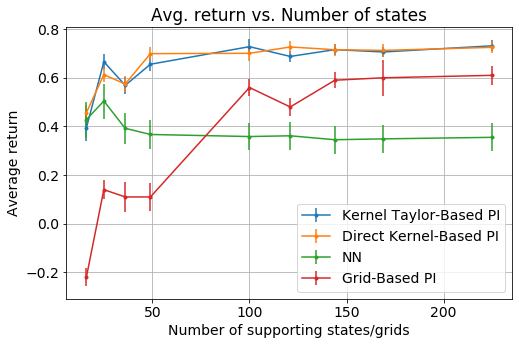}
    \caption
    {\small 
        The comparison of the average return of the policies computed from the four algorithms. 
        The x-axis is the number of supporting states/grids used in computing the policy.
        The y-axis shows the average return.
    }
    \label{fig:plane-nav-performance}
    \vspace{-15pt}
\end{figure}

\begin{figure*}[t]
    \centering
    \subfloat[]{\label{fig:even-support-state}\includegraphics[width=0.225\linewidth, height=1.3in]{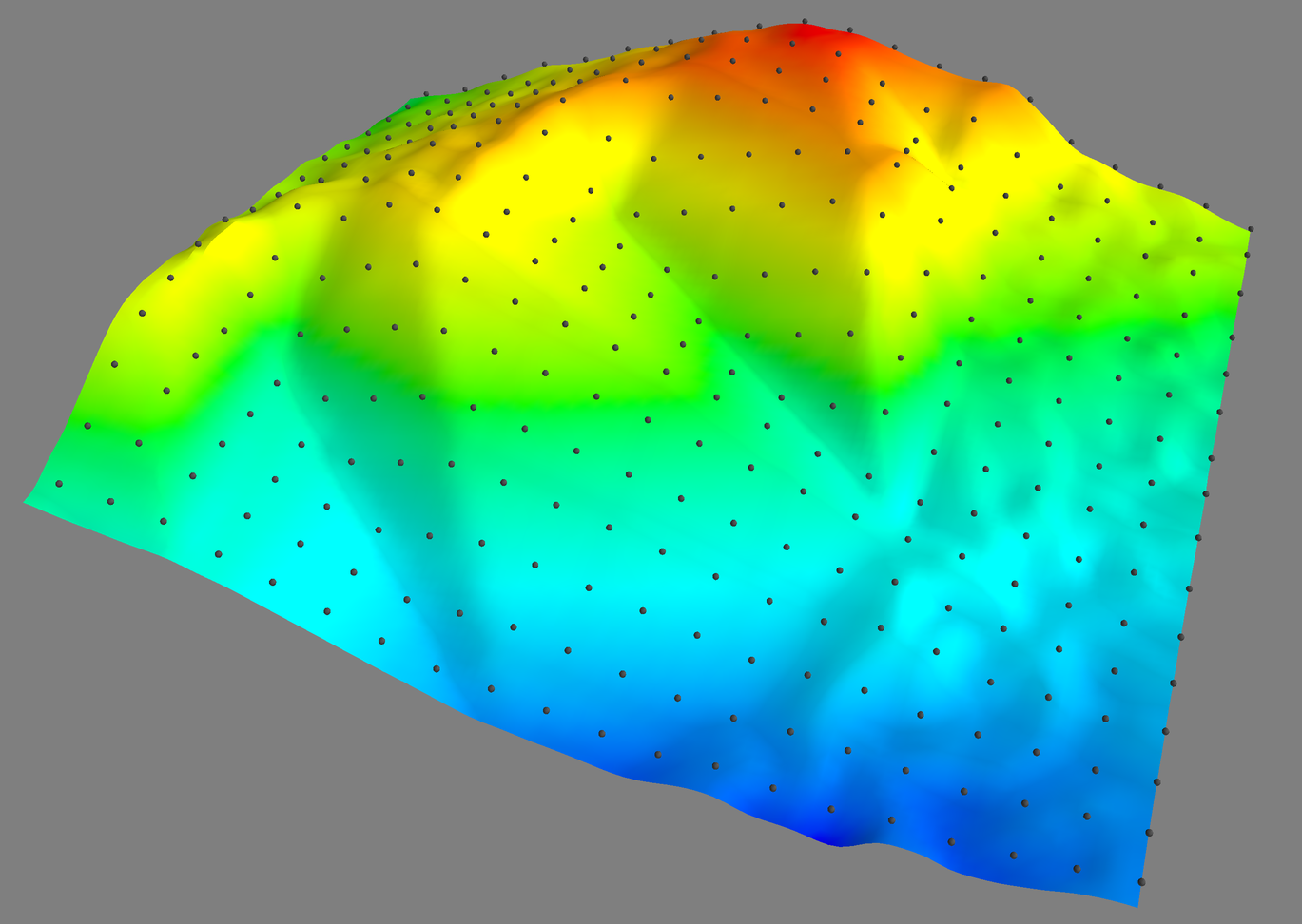}}
    \quad
    \subfloat[]{\label{fig:even-3d-policy}\includegraphics[width=0.235\linewidth,height=1.3in]{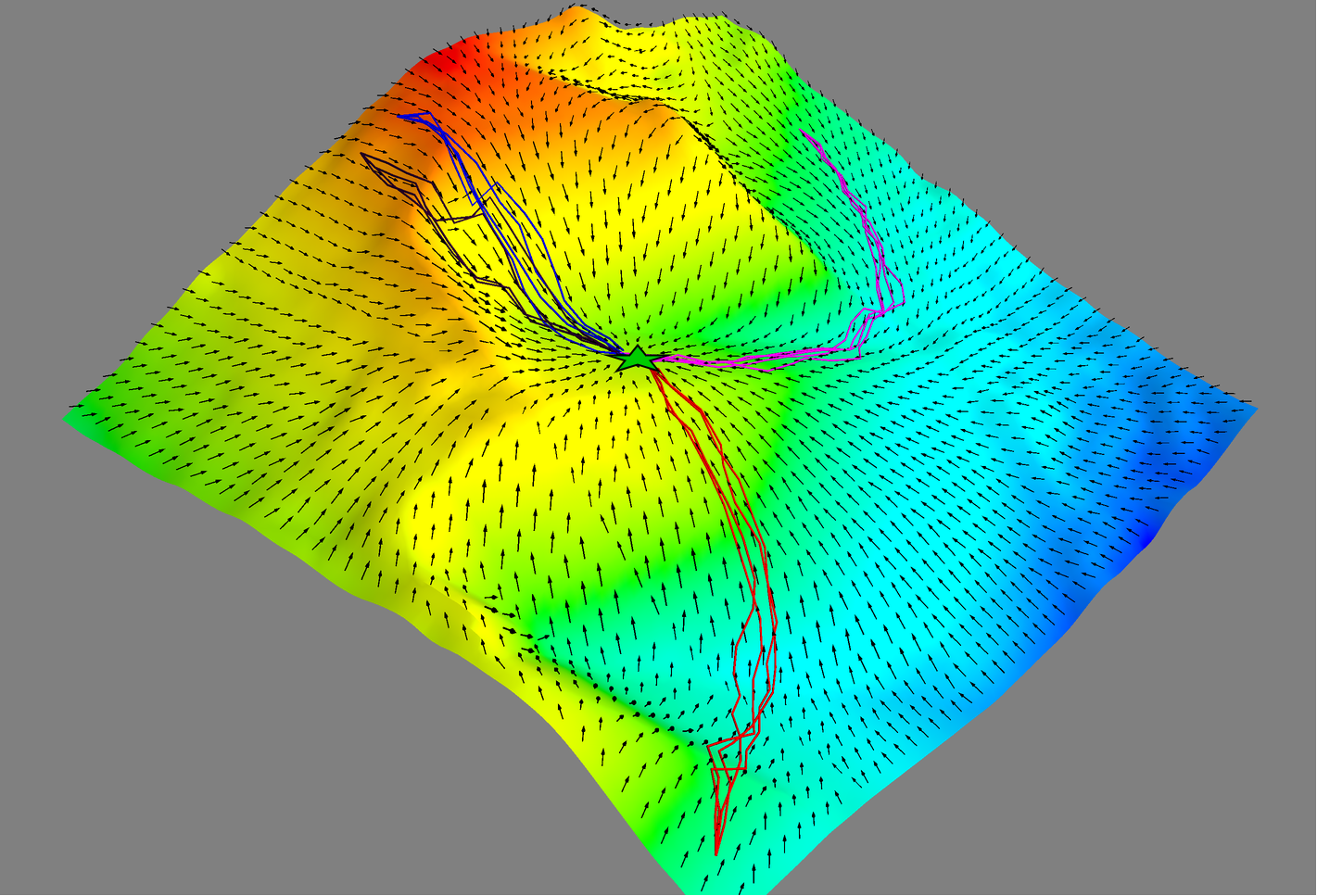}}
    \quad
    \subfloat[]{\label{fig:weighted-support-state}\includegraphics[width=0.23\linewidth,height=1.3in]{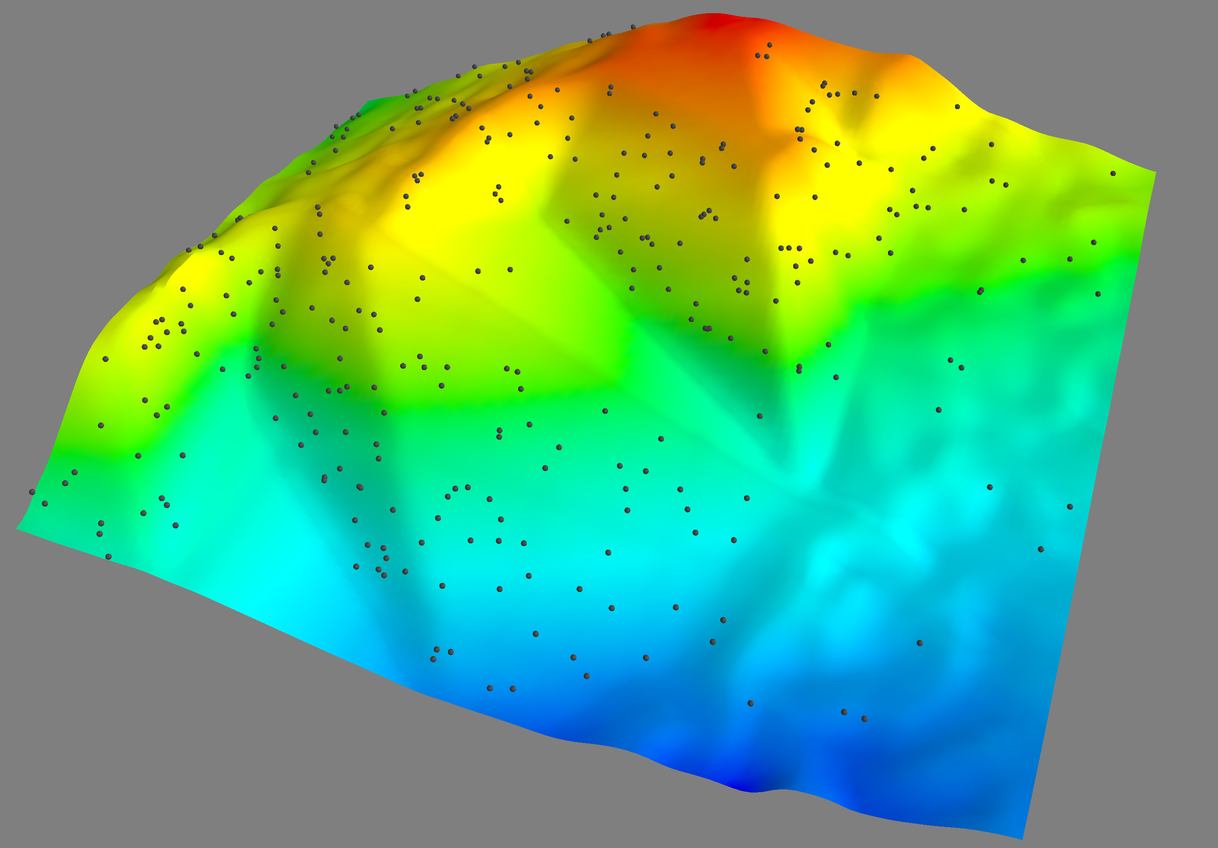}}
    \quad
    \subfloat[]{\label{fig:weighted-3d-policy}\includegraphics[width=0.239\linewidth,height=1.3in]{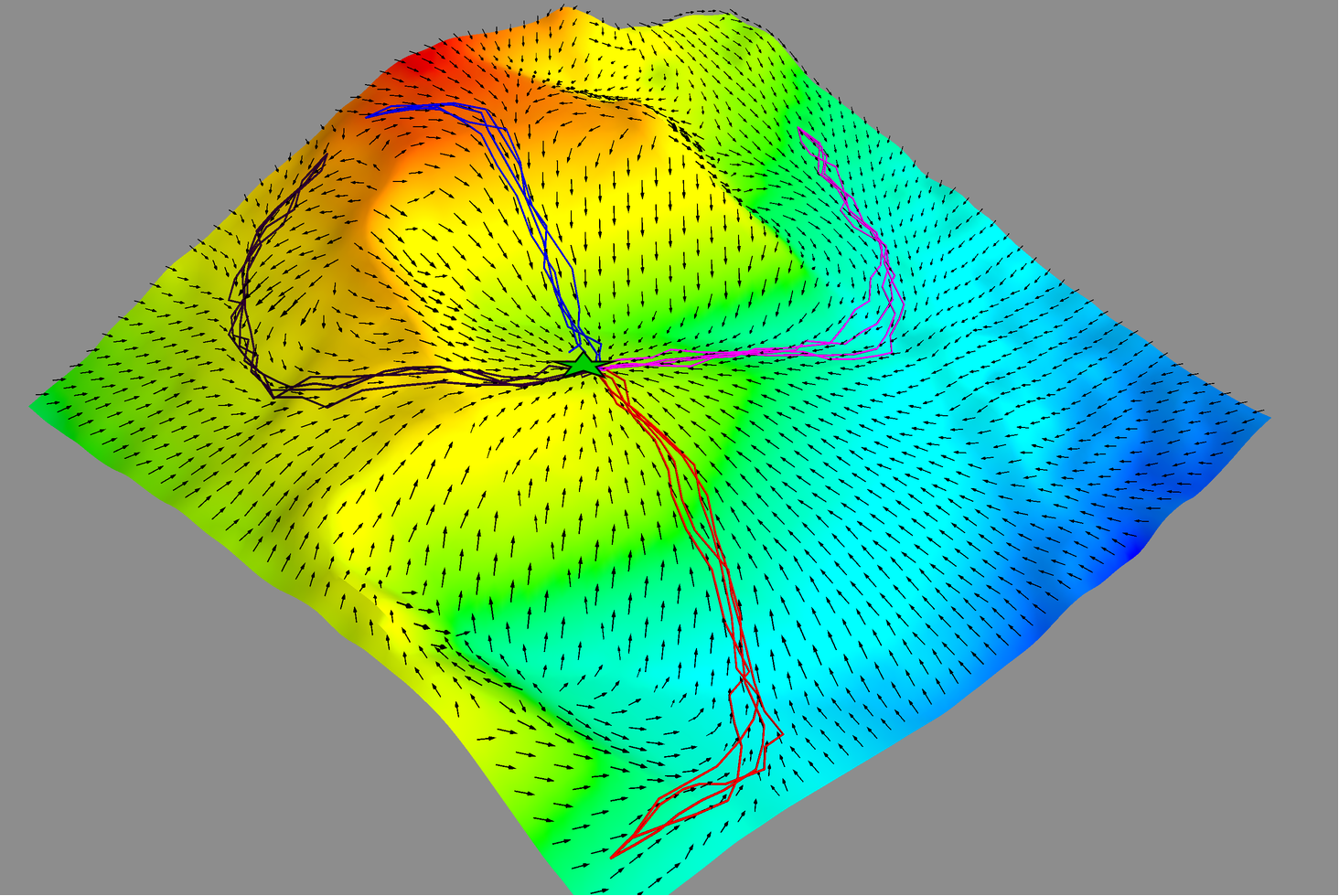}}
    \caption{\small 
    Supporting state distribution and the policy for evenly-spaced selection and importance sampling-based selection.
    The 3D surface shows the Mars digital terrain model obtained from HiRISE. 
    Supporting states and the policies are shown in black dots and vector fields, respectively.
    The colored lines represent the sampled trajectories, which initiate from four different starting positions.
    (a)(b) The evenly-spaced supporting states and the corresponding policy and trajectories; 
    (c)(d) The supporting states generated by importance sampling and the corresponding policy and trajectories.   
    }
    \label{fig:policy-3d} \vspace{-10pt}
\end{figure*}
To evaluate the effect of supporting states, 
we place evenly-spaced supporting states (in a lattice pattern) with different spacing resolution. 
Besides the number of states, the kernel lengthscale (see Appendix~\ref{Gaussian-kernel}) and the regularization parameter $\lambda$ are the other two {\em hyperparameters} governing the performance of our algorithm. 
We present the grid-based hyperparameter search results using four different configurations of supporting states shown in Fig.~\ref{fig:hyper-parameter}.
The lengthscale and regularization parameters are searched over the same values, $\{0.5, 1, 1.5, 2, 2.5, 3\}$.
By entry-wise comparison among the four matrices in Fig.~\ref{fig:hyper-parameter}, we can observe that increasing the number of states leads to improving performance in general. 
However, we can find that the best performed policy is given by the $10\times 10$ supporting states configuration (Fig.~\subref*{fig:taylor-pi-100}) which is not the scenario with the best spacing resolution. 
This indicates that {\em a larger number of states can also result in a deteriorating solution}, and the performance of the algorithm is a matter of {\em how the supporting states are placed (distributed)}, instead of {\em the number (resolution) of state discretization}.  
Furthermore, we can gain some insights on how to select the hyperparameters based on the number of supporting states.
Low-performing entries (highlighted with red) occur more often on the left side of the performance matrix when the number of supporting states increases.
It implies that with more supporting states, the algorithm requires a stronger regularization (i.e., greater $\lambda$ described in Section~\ref{kernel-taylored-P-E}).
On the other hand, high-performing policies (indicated by yellow) appear more on the bottom of the performance matrix when a greater number of supporting states  present,
which means that a smaller length scale is generally required given a larger quantity of supporting states.

We further compare our kernelized value function representation against other three  
variants of the value function approximations. 
Specifically, the first one is the direct kernel-based approximation method using a Gaussian kernel.
This method is similar to the one in~\cite{kuss2004gaussian}, but with a fully known transition function.
The second one uses neural networks (NNs) as the value function approximator.
We setup the NN configuration similar to a recent work~\cite{heess2015learning}.
In detail, we use a shallow two-layer network with 100 hidden units in each layer.
Its parameters are optimized through minimizing the squared Bellman error via gradient descent. 
Since there is no closed-form solution to compute the expected next state value when using NN as a function approximator, we use Monte-Carlo sampling to estimate the expected value at the next states $\mathbb{E}[v^{\pi}(s')|s]$. 
The third method is the grid-based approximation method.
It first transforms the continuous MDP to a discrete version, where each state is regarded as a grid.
Then, it uses the vanilla policy iteration to solve the discretized MDP. 

The comparison among above methods aims at investigating two important questions: 
\begin{enumerate}
    \item How does the kernelized value function representation compare to other representations (NN and grid-based) in terms of the final policy performance? 
    \item Compared to our method, the direct kernel-based method not only requires the fully known transition function, but also restricts the transition to be a Gaussian distribution.
    Can our method with only mean and variance obtain similar performance as the direct kernel-based method?
\end{enumerate}
To answer these questions, we choose the transition function as a Gaussian distribution.
Its mean is the selected next waypoint. 
We set the standard deviation of the transition function to be $0.2m$ on both axes during the experiment.
The transition probability models the accuracy of the low-level motion controller: more accurate controller leads to smaller uncertainty. 
To perform fair comparisons, we use the same supporting states and apply hyperparameter search to all the methods.


The results for four methods are shown in Fig.~\ref{fig:plane-nav-performance} in terms of the average return.
The first question is answered by the fact that the kernel-based methods (kernel Taylor-based and direct kernel-based PI) consistently outperforms the other two methods. 
Moreover, 
our method has the performance as good as the direct kernel-based method which however requires the prerequisite full distribution information of the transition. 
This indicates that our method can be applied to broader applications that do not have full knowledge of transition functions.
In contrast to the grid-based PI, the kernel-based algorithms and NN can achieve moderate performance even with 
a small number of supporting states.  
It indicates that the continuous representation of the value function is crucial when supporting states are sparse. 
However, increasing the number of states does not improve the performance of the NN.

\begin{figure}[t] 
    \centering
    \subfloat[]{\label{fig:computation-time}\includegraphics[width=0.5\linewidth,height=1.45in]{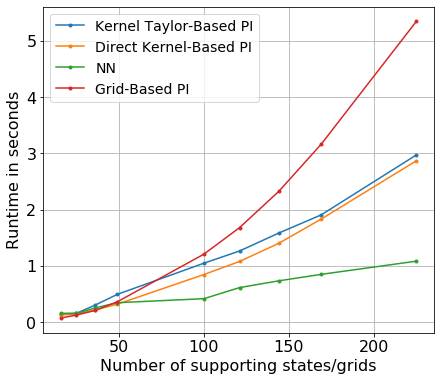}}
    \subfloat[]{\label{fig:convergence-iter}\includegraphics[width=0.5\linewidth,height=1.45in]{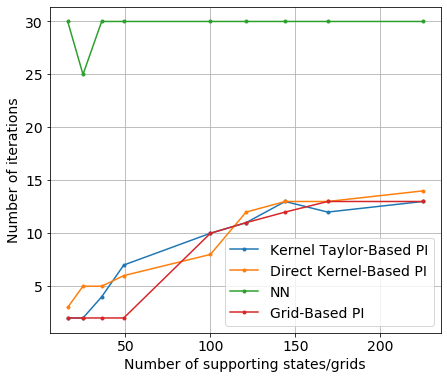}}
    \caption
    {\small 
        Computational time comparisons of the four algorithms with changing number of states. 
        (a) The computational time per iteration.
        (b) Number of iterations to convergence.
    }
    \label{fig:computation-statistics}
    \vspace{-10pt}
\end{figure}

In Fig.~\ref{fig:computation-statistics}, we compare the computational time and the number of iterations to convergence. 
The computational time of our method is less than the grid-based method as revealed in Fig.~\subref*{fig:computation-time}. 
We notice that there is a negligible computational time difference between our method and the direct method.
As a parametric method, NN has the least computational time, and only linearly increases, but it does not converge as indicated by Fig.~\subref*{fig:convergence-iter}. 

The function values and the final policies are visualized in Fig.~\ref{fig:value-fn-policy}. 
All the methods except for the NN obtain reasonable approximations to the optimal value function.
Compared to our method, the values generated by the grid-based method are discrete ``color blocks", thus the obtained policy is non-smooth.
The direct kernel-based method obtains a slightly more ``aggressive" (dangerous) policy in contrast to our method.

\subsection{Martian Terrain Navigation}
In this experiment, we consider the autonomous navigation task on the surface of Mars with a rover.
We obtain the Mars terrain data from \textit{High Resolution Imaging Science Experiment} (HiRISE)~\cite{mcewen2007mars}.
Since there is no explicitly presented ``obstacle", the robot only receives the reward when it reaches the goal.
If the rover attempts to move on a steep slope, it may be damaged and trapped within the same state with probability proportional to the slope angle.
Otherwise, its next state is distributed around the desired waypoint specified by the current action.
This indicates that the underlying transition function should be the mixture of these two situations. 
It is reasonable to assume that the means of the two cases are given by the current state and the next waypoint, respectively. 
We can similarly have an estimate of the variances.
The mean and the variance of the transition function can then be computed using the law of total expectation and total variance, respectively.

Due to the complex and unstructured terrestrial features, evenly-spaced supporting state points may fail to best characterize the underlying value function.
Also, to keep the computational time at a reasonable amount while maintaining a good performance, we leverage the importance sampling technique to sample the supporting states that concentrate around the dangerous regions where there are steep slopes.
This is obtained by first drawing a large number of states uniformly covering the whole workspace.
For each sampled state, we then assign a weight proportional to its slope angle.
Finally, we resample supporting states based on the weights. 
To guarantee the goal state to have a value, we always place one supporting state at the center of the goal area.

\begin{figure}[t]
    \centering
    \subfloat[]{\label{fig:terrain-policy-map-weighted-2d}\includegraphics[height=1.4in]{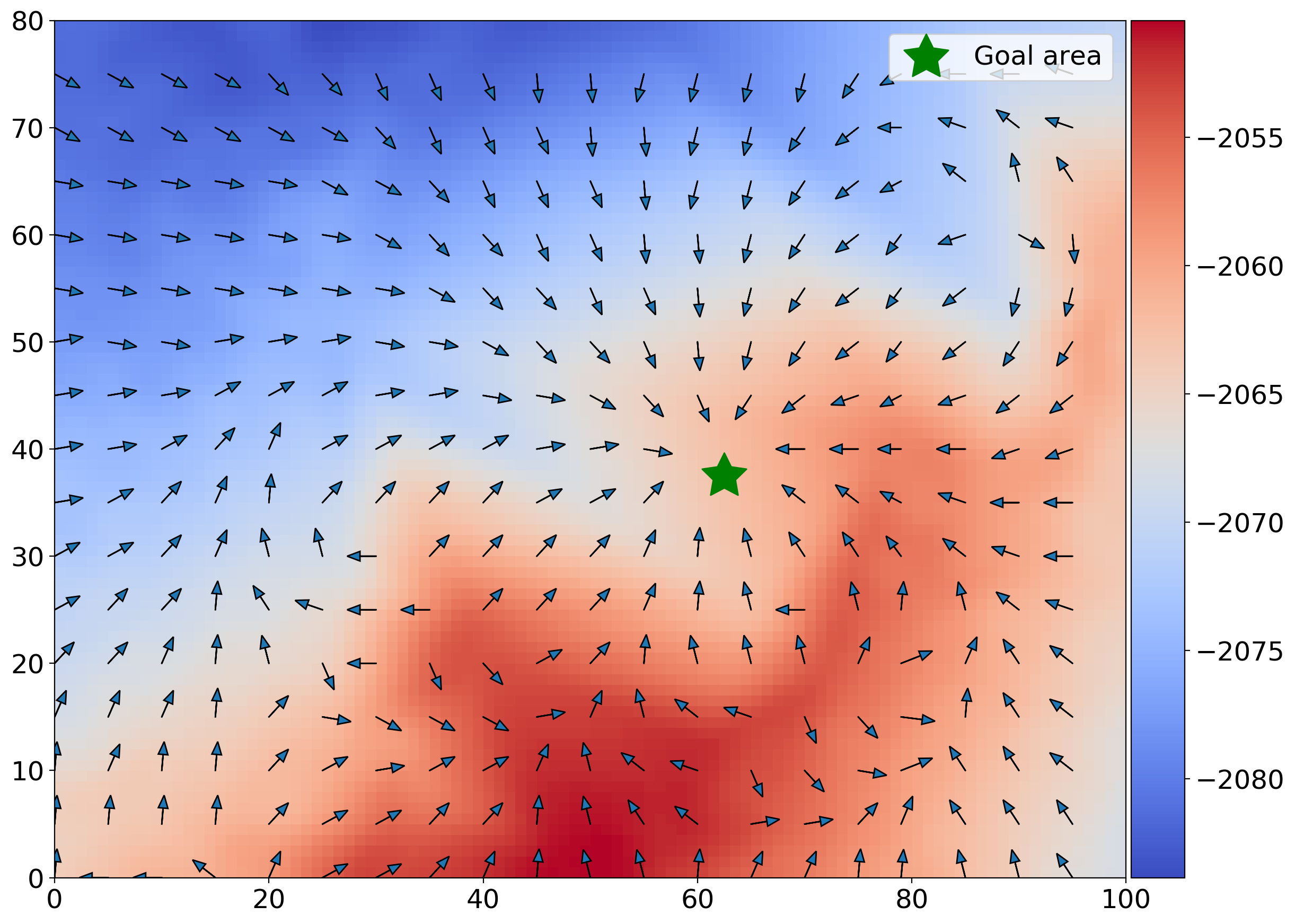}}
   \ \ 
    \subfloat[]{\label{fig:bar-chart}\includegraphics[height=1.4in]{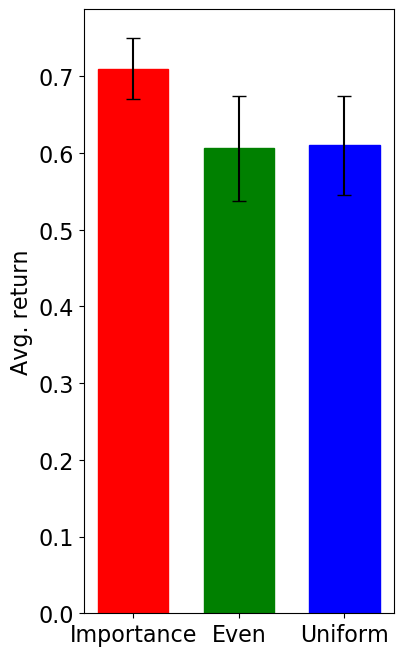}}
    \caption{
    \small 
    (a) The top-down view of the  Mars terrain surface as well as the policy generated by our method with the importance sampling selection. The colormap indicates the height (in meters) of the terrain.
    (b) The comparison of average return among three supporting state selection methods using the same number of states. Red, green, and blue bars indicate the performance of importance sampling selection, evenly-spaced selection, and uniform distribution sampling selection, respectively. 
    }
    \label{fig:policy-2d} \vspace{-10pt}
\end{figure}

Fig.~\subref*{fig:even-support-state} and Fig.~\subref*{fig:weighted-support-state} compare the two methods for supporting state selections.
The supporting states given by the importance sampling-based method are dense around the slopes.
These supporting states better characterize the potentially high-cost and dangerous areas than the evenly-spaced selection scheme.  
We selected four starting locations from where the rover needs to plan paths to arrive at a goal location. 
For each starting location, we conducted multiple trials following the produced optimal policies.
The 
trajectories generated with the importance sampling states 
in Fig.~\subref*{fig:weighted-3d-policy} attempt to approach the goal (green star) with minimum distances, and at the same time, avoid high-elevation terrains.
In contrast, the trajectories obtained using the evenly-spacing states in Fig.~\subref*{fig:even-3d-policy} approach the goal in a  more aggressive manner which can be risky in terms of safety. 
It indicates that a good selection of supporting states can better capture the state value function and thus produce finer solutions. 
This superior performance can also be reflected in Fig.~\subref*{fig:bar-chart}.
The policy obtained by the uniformly sampled states shows similar performance to the one generated by the evenly-spacing states, both of which yield smaller average return than the importance-sampled case. 
A top-down view of the policy is shown in Fig.~\subref*{fig:terrain-policy-map-weighted-2d} where the background colormap denotes the elevation of terrain.

\section{Conclusion}
This paper presents an efficient policy iteration algorithm to solve the continuous-state Markov Decision Process by integrating the kernel value function representation and the Taylor-based approximation to Bellman optimality equation. 
Our algorithm alleviates the need for heavily searching in continuous state space and the need for precisely modeling the state transition functions. 
We have thoroughly evaluated the proposed method through simulations in both simplified and realistic planning scenarios.  
The experiments comparing with other baseline approaches show that our proposed framework is powerful and flexible,  and the performance statistics reveal superior efficiency and accuracy of our algorithms. 



\appendix
\subsection{Gaussian Kernel for kernel Taylor-Based Approximate Policy Evaluation}\label{Gaussian-kernel}
Because Gaussian kernels are widely used in the studies of kernel methods, this section presents the necessary 
derivations to aid the application of Gaussian kernels to our proposed kernel Taylor-Based approximate methods.

Gaussian kernel functions on states $s'$ and $s$ have the form $k(s', s) = c\times\exp\left((-\frac{1}{2}(s'-s)^T\Sigma^{-1}_s(s'-s)\right)$, where $c$ is a constant and $\Sigma_s$ is a covariance matrix. Note that $\Sigma_s$ is referred to as the length-scale parameter in our paper. Due to limited space, we only provide formula below for the first and second derivatives of the Gaussian kernel functions. These formula are necessary when Gaussian kernels are employed (for example, Eq.\eqref{equations-for-M}). In fact, we have 
\begin{equation}\label{Gaussian-first-dev}
    \nabla_{s'}k(s', s) = -\Sigma_1^{-1}(s'-s)k(s', s),
\end{equation}
and
\begin{equation}
\begin{split}
   & \nabla_{s'}\cdot \sigma_s\nabla_{s'} k(s', s) = - tr(\sigma_s\,\Sigma_s^{-1})\,k(s', s)\\ 
   &+(s'-s)^T\Sigma_s^{-T}\sigma_s\,\Sigma_s^{-1}(s'-s)k(s', s),
\end{split}
\end{equation}
where $tr(\cdot)$ denotes the trace of the matrix.

\newpage
\bibliographystyle{plainnat}
\bibliography{ref}

\end{document}